\documentclass[runningheads]{llncs}
\usepackage[T1]{fontenc}
\usepackage{graphicx,verbatim}
\usepackage{amsmath}
\usepackage{amsfonts}
\usepackage{amssymb}
\usepackage{mathtools}
\usepackage{booktabs}
\usepackage{enumitem}
\usepackage{hyperref}       
\usepackage[table]{xcolor}

\hypersetup{
    colorlinks,
    linkcolor={red!50!black},
    citecolor={blue!50!black},
    urlcolor={blue!50!black}
}

\begin{document}
\title{Latent-to-Latent Flow for\\Volumetric Stochastic Segmentation}
\titlerunning{Latent-to-Latent Flow for Volumetric Stochastic Segmentation}
%

\author{Omar Todd \inst{1} \and Sooha Kim \inst{2} \and Raghav Mehta \inst{1} \and Katherine Mackay \inst{3} \and David Bernstein \inst{2} \and
Alexandra Taylor \inst{2} \and Fabio De Sousa Ribeiro \inst{1,\text{$\dagger$}} \and Ben Glocker\inst{1,\text{$\dagger$}}}
\begingroup
\renewcommand{\thefootnote}{\ensuremath{\dagger}}%
\footnotetext{Joint senior authors.}%
\endgroup
\authorrunning{Todd et al.}
\institute{Imperial College London, UK \and The Royal Marsden Hospital, London, UK \and Imperial College Healthcare NHS Trust, London, UK\\
\email{omar.todd16@imperial.ac.uk}}

  
\maketitle              

\begin{abstract}
Uncertainty arising from inter-observer variability in medical image segmentation plays an important role in developing treatment plans. Research in this area is inhibited by the lack of multiple annotations for large-scale medical datasets, especially for volumetric data, which suffers from additional scaling and computational complexity challenges. Flow matching has emerged as a powerful framework for generative modelling and has also been demonstrated to maintain strong performance when working with latent representations of images. In this work, we introduce a latent-to-latent flow technique for stochastic segmentation of medical volumes via encoded representations of both the image and label space. We evaluate our method on two challenging applications covering delineation uncertainty for radiotherapy planning and multiple organ structure segmentation, improving efficiency up to $14\times$ compared with full resolution models while maintaining clinically relevant performance. Code available at: \href{https://github.com/biomedia-mira/L2L-Flow}{https://github.com/biomedia-mira/L2L-Flow}.


\end{abstract}

\section{Introduction}
Medical image segmentation aims to provide accurate contouring of anatomical structures to aid clinicians in developing treatment plans. Segmenting medical images is an inherently ambiguous task due to uncertainty in the exact positions of organ and tissue boundaries. 
Incorporating uncertainty into the planning process is important for improving patient outcomes \cite{mccrindle2021radiology,kinoshita2025quantification}. Uncertainty in machine learning is broadly categorized into two classes: epistemic, which concerns the model, and aleatoric, which concerns the data \cite{zou2023review}. Inter-observer variation (IOV) \cite{mcerlean2013intra}, arising from clinical experts contouring a structure differently, is an important source of aleatoric uncertainty that deterministic segmentation models, such as U-Net  \cite{ronneberger2015u}, are unable to accurately capture. This requires the use of stochastic segmentation models, which learn a probability distribution over the plausible segmentation maps to simulate clinically meaningful uncertainty.

While large amounts of medical imaging data are represented as 3D volumes, much of the stochastic segmentation research and benchmarking is developed for 2D datasets~\cite{armato2011lung,cordts2016cityscapes}. Data and label scarcity is a prevalent issue due to the labour-intensive nature of collecting multiple expert annotations, which is exacerbated for 3D volumes. Additionally, probabilistic modelling of volumetric annotations comes with other challenges such as increased computational complexity. This can be an issue for time-sensitive applications such as adaptive radiotherapy planning. Recently, flow matching has emerged as an attractive framework for generative modelling on natural images \cite{lipmanflow,liu2023flow,albergo2023building}; however, its suitability for volumetric stochastic segmentation has not been explored and is the main focus of this work. Our primary contributions are as follows:

\begin{enumerate}[label=(\roman*)]
\item We identify performance degradation of Flow-SSNs~\cite{de2025flow} on high-resolution volumetric medical data, and introduce a bespoke time-shifted noise schedule resulting in significantly improved stability and performance, setting a new benchmark for volumetric stochastic segmentation.
\item We introduce the Latent-to-Latent Flow (L2L-Flow), an efficient and scalable model alternative to our time-shifted Flow-SSN, allowing for a wider range of volumetric clinical applications. Notably, L2L-Flow provides a controllable efficiency/performance trade-off, demonstrating up to 14$\times$ faster inference compared to Flow-SSNs whilst retaining highly competitive performance.
\item We evaluate our flow models on two challenging medical datasets for radiotherapy planning and multi-organ segmentation, demonstrating clinically relevant uncertainty estimation capabilities, and highlighting the potential of generative approaches for volumetric segmentation under label ambiguity.
\end{enumerate}

\section{Related Work}
Probabilistic U-Net~\cite{kohl2018probabilistic} utilized a U-Net alongside a conditional variational auto-encoder to learn a conditional density function for generating stochastic segmentations.  PhiSeg~\cite{baumgartner2019phiseg} built on this model by introducing a hierarchical probabilistic model for segmenting the image via latent distributions at different resolutions. Stochastic Segmentation Networks (SSNs)~\cite{monteiro2020stochastic} explored limitations of the pixel-wise independence assumed in previous work, and employed a more expressive low-rank Gaussian distribution for modelling pixel variance. This model, however, is prone to stability issues during training, and the low-rank approximation imposes limitations on the complexity of the distribution it can learn \cite{de2025flow}. More recently, Mixture-of-Stochastic-Experts (MoSE) modelled each annotator as a separate distribution, from which samples were optimized via an optimal transport loss function \cite{gaomodeling}. For 3D data, this model can suffer from scaling issues if it is desired to have a high number of samples per mode during training.

Diffusion models are another generative model class that have been successfully applied to the task of stochastic segmentation in 2D \cite{wolleb2022diffusion,zbinden2023stochastic}. While they have demonstrated the ability to generate high-quality segmentations, Denoising Diffusion Probabilistic Model (DDPM) \cite{ho2020denoising} based methods are very expensive at inference time, which becomes increasingly undesirable as the data scale increases. More efficient options such as Denoising Diffusion Implicit Models (DDIM) \cite{song2020denoising} can help with this; however, the paths induced by these methods often have a complex shape that is difficult to solve. Most recently, the flow matching paradigm has emerged as a simple yet powerful setup for generative modelling \cite{lipmanflow,liu2023flow,albergo2023building}. Flow-SSNs \cite{de2025flow} introduced a modified rectified flow with a learnable image-conditioned prior in one-hot label space and a categorical likelihood for medical stochastic segmentation. This approach showed strong performance in 2D, while providing vastly superior efficiency compared to diffusion-based stochastic segmentation models \cite{zbinden2023stochastic}. Recently, latent rectified flows~\cite{esser2024scaling} have also gained traction. Ma et al.~\cite{ma2025cardiacflow} applied them to cardiac segmentations to facilitate data augmentation for shape completion. Focusing on stochastic segmentation, however, it is crucial for the latent flow to utilize the rich structural information in the high-resolution images and labels to preserve strong clinical utility when modelling the distribution in a lower-dimensional space.

\section{Methodology}
\subsection{Neural ODEs and Flow Matching}

Continuous Normalising Flows (CNFs)~\cite{chen2018neural} learn a mapping from a source $\mathbf{x}_0 \sim p_{\text{src}}$ to a target distribution $\mathbf{x}_1 \sim p_\text{tgt}$ via an ordinary differential equation (ODE)
\begin{align}
    \label{eq: ode}
    \mathrm{d}\mathbf{x}_t = v_t(\mathbf{x}_t;\theta) \, \mathrm{d}t, \qquad t \in [0,1],
\end{align}
where $v_t$ is a time-dependent velocity field parameterised by a neural network. 

Flow Matching~\cite{lipmanflow,liu2023flow,albergo2023building} trains CNFs without simulating trajectories by specifying a conditional probability path between $\mathbf{x}_0$ and $\mathbf{x}_1$, such as simple linear interpolation $\mathbf{x}_t = (1-t)\mathbf{x}_0 + t\mathbf{x}_1$, yielding a least-squares regression objective:
\begin{align}
\label{eq: fmloss}
\mathcal{L}_{\mathrm{FM}}(\theta)
&=
\mathbb{E}_{t,\mathbf{x}_0,\mathbf{x}_1}
\left[
\left\|
v_t(\mathbf{x}_t;\theta) - v_t^\star(\mathbf{x}_t \mid \mathbf{x}_1)
\right\|^2
\right],
\end{align}
where $t \sim \mathcal{U}[0,1]$, $\mathbf{x}_0 \sim p_{\text{src}}$ and $\mathbf{x}_1 \sim p_{\text{tgt}}$. Flow-SSNs~\cite{de2025flow} learn a mapping from an image conditional prior $\mathbf{u}|\mathbf{x} \sim p_{\text{src}}$ to a label map distribution $\mathbf{y} \sim p_{\text{tgt}}$. Since targets are discrete, Flow-SSNs use a linear interpolant in one-hot space $\mathbf{y}_t = (1-t)\mathbf{u} + t\mathbf{y}$, and an expected categorical likelihood: $
\max_{\lambda,\theta} \mathbb{E}_{t,\mathbf{u},\mathbf{y}}
\left[\log p(\mathbf{y} \mid \mathbf{y}_t;\theta)
\right]$, where the source $\mathbf{u} \sim p_{\text{src}}(\mathbf{u}|\mathbf{x};\lambda)$ is learnt from data rather than being fixed.

\subsection{Time-Shifted Noise Scheduling}
Recent work on modern flows showed that the use of a shifted noise schedule leads to improved sample quality for high-resolution image data~\cite{labs2025flux}. 
We find that careful calibration of the noise schedule is critical to the effective use of Flow-SSNs in high-dimensional volumetric segmentation tasks, which has been previously overlooked~\cite{de2025flow}.
We propose a modified version of the shift schedule~\cite{labs2025flux}, which accounts for our path direction going from $t=0$ to $t = 1$, yielding:
\begin{align}
&&\mathbf{y}_t = (1-t_s)\mathbf{u} + t_s\mathbf{y},  && \text{where} && t_s = \frac{t}{\alpha - t(\alpha - 1)}.&&
\end{align}
where higher values of the hyperparameter $\alpha$ induce more aggressive noising. As will be demonstrated, this simple modification unlocks significant training stability and performance improvements for Flow-SSNs in volumetric settings.
\begin{figure}[t]
\centering
\includegraphics[trim={0pt 16pt 0pt 17pt},clip,width=.9\textwidth]{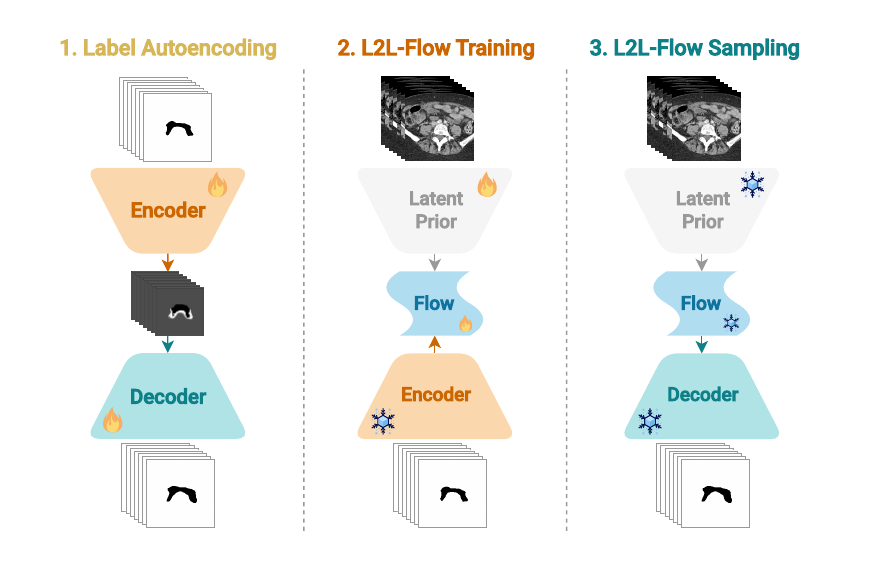}
\caption{
\textbf{Architectural overview of L2L-Flow}.
(\textit{Left}) A volumetric segmentation label autoencoder is trained to obtain latent label representations. (\textit{Middle}) A latent flow is then trained to learn a transport from a conditional prior given input volumes to frozen latent label maps. (\textit{Right}) Stochastic segmentation is performed by integrating the ODE, mapping conditional samples from the volume prior to latent label predictions. Finally, the ODE solution is upsampled to full resolution using the label decoder.
} 
\label{architecture}
\end{figure}
\subsection{Latent-to-Latent Flow}
Unlike standard latent flows/diffusions which model the distribution of a single high-dimensional image variable $p(\mathbf{x})$, generative segmentation seeks to estimate the label conditional distribution $p(\mathbf{y}\mid \mathbf{x}) =\int p(\mathbf{y}, \mathbf{z} \mid \mathbf{x}) \, \mathrm{d}\mathbf{z}$, where both $\mathbf{x}$ and $\mathbf{y}$ are high-dimensional, and $\mathbf{z}$ is a latent variable. Flow-SSNs~\cite{de2025flow} model this joint in observation space, making sampling too computationally expensive for volumes. To overcome this, we propose Latent-to-Latent Flow (L2L-Flow), a new scalable approach that learns a flow transport from a \textit{latent} volume-conditional prior to a \textit{latent} label distribution. As described in Fig.~\ref{architecture}, the first stage consists of training a volumetric segmentation label autoencoder $D_\psi(E_\phi(\mathbf{y}))$ to obtain compressed label latents $\mathbf{z}_1 = E_\phi(\mathbf{y})$. We then define the latent linear flow interpolant:
\begin{align}
    \mathbf{z}_t = (1-t) E_\lambda(\mathbf{x}) + t E_\phi(\mathbf{y}), \qquad t \in [0,1],
\end{align}
where \textit{both} the source and target distributions are parameterised by encoders:
\begin{align}
    && p_{\text{src}}(\mathbf{z}_0 \mid \mathbf{x};\lambda)= \mathcal{N}(E_\lambda(\mathbf{x}), I), && \text{and} && p_{\text{tgt}}(\mathbf{z}_1 \mid \mathbf{y};\phi) = \delta(\mathbf{z}_1 - E_\phi (\mathbf{y})).&&
\end{align}
Dissimilar to Flow-SSNs, the latent target variable $\mathbf{z}_1 = E_\phi(\mathbf{y})$ here is no longer a discrete one-hot label map, so a least-squares regression objective can be used:
\begin{align}
\label{eq: fmloss2}
\min_{\lambda,\theta} \mathbb{E}_{t,\mathbf{z}_0\sim p_{\text{src}},\mathbf{z}_1\sim p_{\text{tgt}}}
\left[
\left\|
v_t(\mathbf{z}_t;\theta) - (E_\phi (\mathbf{y}) - \mathbf{z}_0)
\right\|^2
\right],
\end{align}
noting that $v_t(\mathbf{z}_t;\theta)$ and $E_\lambda (\mathbf{x})$ are trained simultaneously, while we find that pretraining and freezing $E_\phi (\mathbf{y})$ leads to improved stability and performance.

At inference time, stochastic segmentation is performed by: \text{(i)} sampling from the latent volume-conditional prior $\mathbf{z}_0 \sim p_{\text{src}}(\mathbf{z}_0 \mid \mathbf{x};\lambda)$; \text{(ii)} solving the flow ODE by integrating $\mathbf{z}_1 = \mathbf{z}_0 + \int_0^1 v_t(\mathbf{z}_t;\theta) \, \mathrm{d}t $; and \text{(iii)} upsampling the ODE solution using the frozen pretrained label decoder $\mathbf{y} = D_\psi(\mathbf{z}_1)$.
As we will demonstrate empirically, L2L-Flow accelerates flow sampling by up to 14$\times$ while maintaining volumetric segmentation performance comparable to full-resolution models.
\section{Experiments}
\subsection{Datasets}
To address the challenges of gathering high-quality IOV for medical data, recent works have researched more efficient annotation schemes using clinician-defined ranges (CDR) \cite{bernstein2021new,mackay20251621,todd2025delineation}. These ranges provide soft inner and outer boundaries within which the true contour is expected to be positioned, while indicating the non-uniform, spatially varying uncertainty of the contour shape. To explore the suitability of CDRs for stochastic segmentation, we evaluate the clinical relevance of samples from probabilistic models trained on our private CT radiotherapy dataset of clinical target volumes (CTV). We also evaluate our method on the publicly available multi-class CURVAS challenge dataset with three annotations from multiple raters for three structures: pancreas, kidney, and liver \cite{riera2025calibration}.

\subsection{Implementation Details}
The radiotherapy and CURVAS datasets used for evaluation in this work had a limited number of cases (55 and 90 cases, respectively). As such, we utilize a 5-fold cross-validation strategy for evaluation. We use the standard generalised energy distance (GED) and diversity\footnote{Recall that diversity is only relevant contextually, as high diversity can be trivially achieved with random noise as a model.} with 16 samples to measure the performance of stochastic segmentation methods~\cite{kohl2018probabilistic,monteiro2020stochastic,de2025flow,gaomodeling}.

\begin{figure}[t]
\centering
\includegraphics[trim={5pt 12pt 5pt 5pt},clip,width=\textwidth]{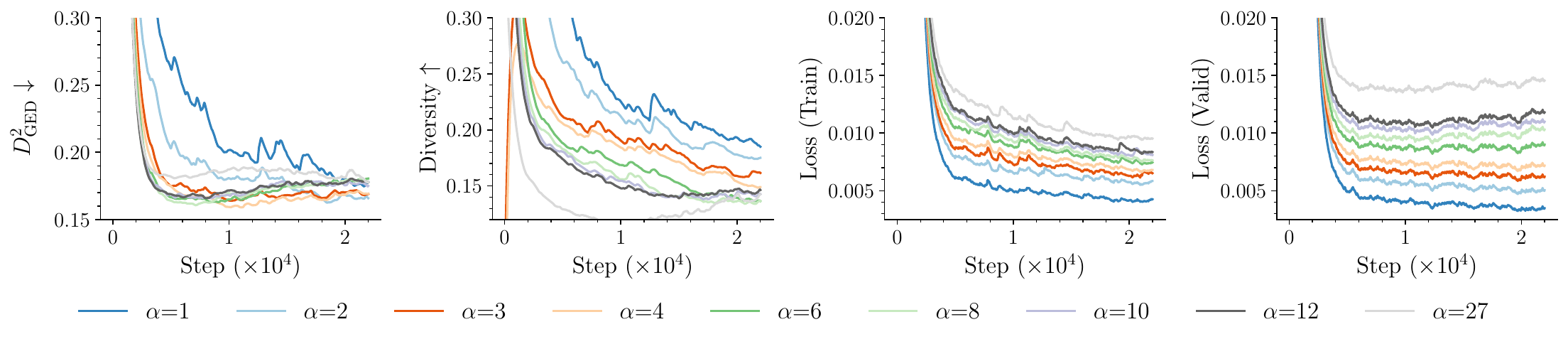}
\caption{\textbf{Ablation of shifted noise schedules}. In high-dimensional volumetric settings, we observe significant performance improvements to GED by shifting the Flow-SSN noise schedule. Higher values of $\alpha$ are more prone to overfitting during training.} 
\label{alpha_curves}
\end{figure}

For both datasets, the high-resolution volumes of interest were taken as $192^3$ ROI crops of the images. Each model was trained using a batch size of 1 with gradient accumulation of 4 steps to accommodate the computational costs of modelling the distribution on large volumes. The models were all trained using a learning rate of 1e-4 for 2000 epochs. All experiments were performed using an NVIDIA L40 (48GB) GPU. The flow matching components of the models described in this work use Residual U-Net backbones~\cite{de2025flow} and use 8 time steps for ODE solving during inference. For the Flow-SSN using a shifted noise schedule, we found $\alpha{=}3$ (see Fig.~\ref{alpha_curves}) to give the best results in our experiments. For L2L-Flow, on the radiotherapy dataset, we down-sample images by a factor of 4 for a latent resolution of $48^3$. On the CURVAS dataset, we additionally down-sample images by a factor of 2 to a $96^3$ resolution. This is achieved via varying the number of levels in the U-Net and the auto-encoder.

\subsection{Radiotherapy Delineation Uncertainty}

\begin{table}[t]
    \centering
    \caption{\textbf{Quantitative cross-validation results on the Radiotherapy dataset}. L2L-Flow demonstrates highly competitive segmentation performance compared to full resolution models, while being more than 14$\times$ faster than Flow-SSN during inference.
    }
    \setlength{\tabcolsep}{5pt}
    \begin{tabular}{lcccrr}
    \toprule
    \textsc{Method} & $D^2_{\text{GED}} \downarrow$ & Diversity $\uparrow$ & 
    Dice $\uparrow$  & s/Img $\downarrow$ & Params
    \\
    \midrule
    MoSE \cite{gaomodeling}
    & 0.183{\scriptsize $\pm$.008} & 
    0.137{\scriptsize $\pm$.005}
    & 0.893{\scriptsize $\pm$.004} & 0.824{\scriptsize $\pm$.010} & 6.9M \\
    PhiSeg \cite{baumgartner2019phiseg}
    & 0.177{\scriptsize $\pm$.007} & 
    0.122{\scriptsize $\pm$.002}
    & 0.898{\scriptsize $\pm$.003} & 4.895{\scriptsize $\pm$.013} & 63M \\
    SSN \cite{monteiro2020stochastic} & 0.167{\scriptsize $\pm$.005} & 0.160{\scriptsize $\pm$.007} &  
    0.896{\scriptsize $\pm$.003} 
    & 0.622{\scriptsize $\pm$.010} & 7.1M \\ Flow-SSN \cite{de2025flow} & 0.165{\scriptsize $\pm$.008} & 0.168{\scriptsize $\pm$.008}&  0.894{\scriptsize $\pm$.004} &  15.296{\scriptsize $\pm$.013}  & 6.9M \\
    \midrule
    Flow-SSN (shifted) & 0.155{\scriptsize $\pm$.005} & 0.164{\scriptsize $\pm$.008} & 0.899{\scriptsize $\pm$.001}  & 13.471{\scriptsize $\pm$.011} & 6.9M \\
    L2L-Flow  & 0.161{\scriptsize $\pm$.004} & 0.182{\scriptsize $\pm$.008} & 0.896{\scriptsize $\pm$.002}  & 0.936{\scriptsize $\pm$.016}  & 9.3M\\
    \bottomrule
    \end{tabular}
    \label{tab:radiotherapy_table}
\end{table}

\begin{figure}[t]
\centering\includegraphics[width=1\columnwidth]{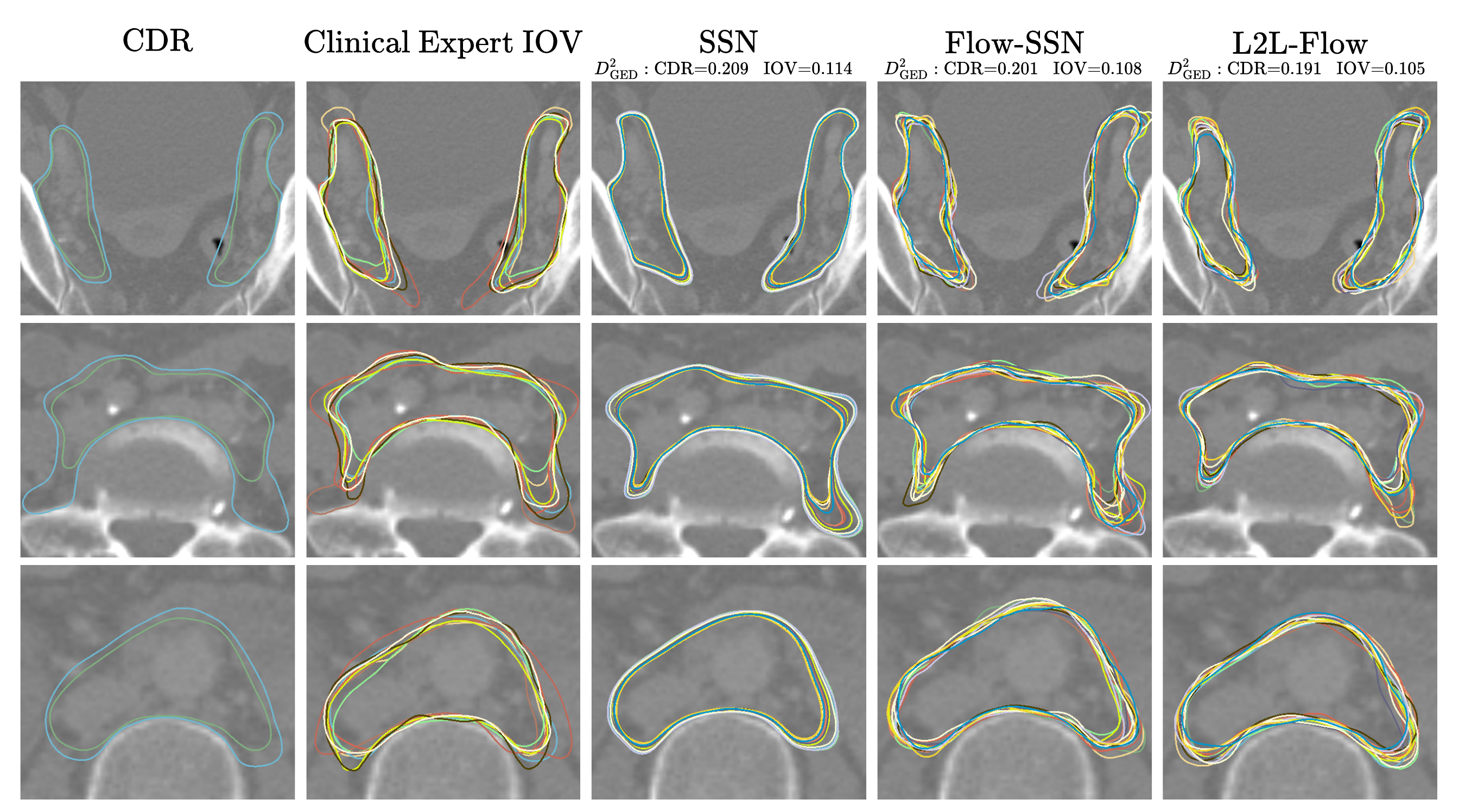}
\caption{\textbf{Qualitative results on the Radiotherapy dataset}. 
Stochastic segmentation samples are compared with a representative example of human expert IOV. 
GED is reported for samples relative to both the CDR and IOV. Crucially, unlike SSN, L2L-Flow and Flow-SSN show clinically relevant variation compared to expert contours.
} 
\label{rad_qual}
\end{figure}

Tab.~\ref{tab:radiotherapy_table} shows the average performance across folds of L2L-Flow alongside each of the baseline methods. For the full-resolution stochastic models, our time-shifted Flow-SSN provides the strongest overall results. We observe that L2L-Flow demonstrates highly competitive performance for GED compared with the full-resolution flow model while being ${\approx}14\times$ more efficient. A qualitative evaluation to better discern the clinical relevance of the best-performing models is shown in Fig.~\ref{rad_qual}. This was performed for a case where ground-truth IOV was available from 10 clinical experts. The expert contours showed numerous overlaps and crossings, which are reflected in the variability of samples from the flow-based models. However, despite strong quantitative performance, the SSN contours only varied in a strong concentric manner. From a clinical standpoint, the strong dissimilarity of these contours to the randomness of human IOV made them unsuitable for downstream planning. This distinction is not captured solely by the GED metric and highlights the need for more holistic comparisons when incorporating machine learning models into the clinical workflow.

\subsection{Multi-Organ Segmentation  (CURVAS Dataset)}
\begin{figure}[t]
\centering
\includegraphics[trim={16pt 0 10pt 0},clip,width=\textwidth]{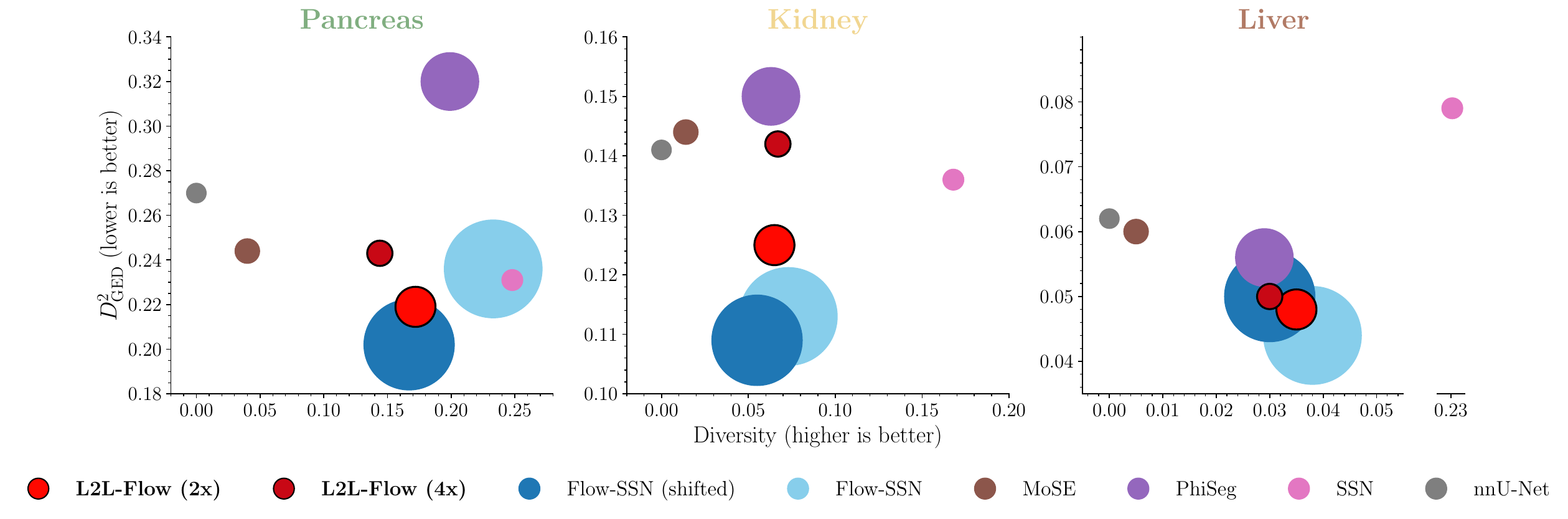}
\caption{\textbf{Quantitative results on the CURVAS dataset}. GED and diversity scores are reported, with circle size proportional to inference time. Flow-SSN (shifted) displays much better performance than the original for the most challenging pancreas class. Reducing the down-sampling used in L2L-Flow from 4$\times$ to 2$\times$ helps bridge the performance gap with full-resolution models, with a trade-off for slower inference time.} 
\label{curvas-quant}
\end{figure}

\begin{figure}[t]
\centering
\includegraphics[width=1\columnwidth]{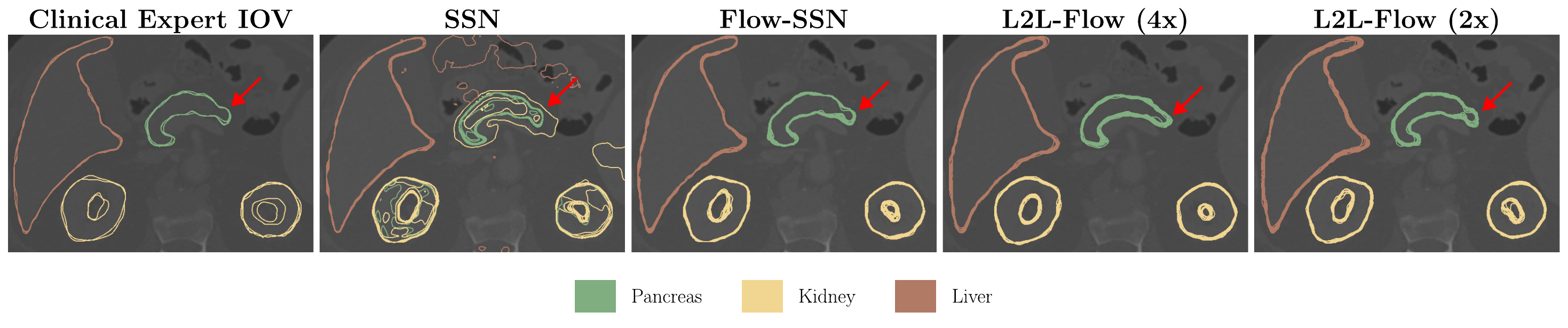} 
\caption{\textbf{Qualitative results on the CURVAS dataset}. SSN produces spurious correlations for multi-class tasks. L2L-Flow at a higher latent resolution can capture more subtle structural properties compared with full-resolution counterparts. Red arrows indicate the 4$\times$ down-sampled model failing to precisely segment a part of the pancreas.} 
\label{curvas-qual}
\end{figure}

 In the original challenge, the best-performing model was the deterministic nnU-Net \cite{isensee2021nnu} which we include as an additional baseline. Per-structure quantitative results are shown in Fig.~\ref{curvas-quant}. Our time-shifted Flow-SSN performs the best, most notably with $3.4\%$ better GED for the pancreas compared to the original Flow-SSN, albeit with substantially slower speeds compared to other models. The more sizeable performance gap on the pancreas between L2L-Flow with 4x downsampling and the best-performing Flow-SSN arises from data compression on an already small structure. To emphasize the flexibility of our model in terms of the efficiency/performance trade-off, we also report results for L2L-Flow at 2x downsampling. This increased performance results from better capturing fine details of the pancreas as shown in Fig.~\ref{curvas-qual}. Despite the increased inference time, this model still remains ${\approx}5\times$ faster than the full resolution Flow-SSN. These experiments also highlighted the non-trivial difficulties of moving existing frameworks to 3D. These challenges can be both methodological, such as SSN's low-rank sampling instability (cf. Fig.~\ref{curvas-qual}), and computational, such as with MoSE's reduced sampling size during training due to memory constraints.

\section{Conclusion}

We introduce L2L-Flow, a scalable generative approach for stochastic segmentation of volumetric data. The proposed model produces clinically meaningful samples while being substantially more efficient in both memory and computation than its full‑resolution counterpart. This is achieved by leveraging high‑level image and label information within a latent rectified flow formulation. Results across multiple datasets demonstrate that extending complex probabilistic models from 2D to 3D is non‑trivial, evidenced by the methodological and training stability issues of prior methods. The radiotherapy experiments, in particular, expose limitations of current stochastic segmentation metrics when used in isolation and underscore the importance of complementing quantitative evaluation with qualitative and clinical considerations when developing machine learning models for clinical workflows. Future work will explore the integration of more expressive image and label encoders, such as those derived from foundation models, to further enhance the capabilities of L2L‑Flow.

\section*{Acknowledgment}
This work was supported by the European Union's Horizon Europe research and innovation programme for the AI-POD project under grant agreement 101080302. Views and opinions expressed are however those of the author(s) only and do not necessarily reflect those of the European Union or HaDEA. Neither the European Union nor the granting authority can be held responsible for them. Additional support was received from the UKRI AI programme, and the EPSRC, for CHAI-EPSRC Causality in Healthcare AI Hub (grant no. EP/Y028856/1) and the Royal Academy of Engineering as part of the Kheiron/RAEng Research Chair. S.K. is supported by the CRUK Convergence Science Centre at The Institute of Cancer Research, London, and Imperial College London (A26234).

\subsection*{Disclosure of interests}
B.G. is a part-time employee of DeepHealth. No other competing interests.

\bibliographystyle{splncs04}
\bibliography{references}
\end{document}